\documentclass{edm_article}

\usepackage[protrusion]{microtype} 
\usepackage{placeins} 
\usepackage{enumitem} 
\setitemize{itemsep=3pt,topsep=3pt}
\usepackage[breaklinks,colorlinks,urlcolor=blue,citecolor=blue]{hyperref} 

\begin{document}

\title{Towards Generalizable Detection of Urgency of~Discussion~Forum Posts}


\numberofauthors{5} 
\author{
\alignauthor
Valdemar Švábenský\\
       \affaddr{University of Pennsylvania}\\
       {\email{\large svabenskyv@gmail.com}}
\alignauthor
Ryan S. Baker\\
       \affaddr{University of Pennsylvania}\\
       \email{\large ryanshaunbaker@gmail.com}
\alignauthor
Andrés Zambrano\\
       \affaddr{University of Pennsylvania}\\
       \email{\large afzambrano97@gmail.com}
\and  
\alignauthor
Yishan Zou\\
       \affaddr{University of Pennsylvania}\\
       \email{\large amieezou@gmail.com}
\alignauthor
Stefan Slater\\
       \affaddr{University of Pennsylvania}\\
       \email{\large slater.research@gmail.com}
}

\toappear{\scriptsize V. Švábenský, R. Baker, A. Zambrano, Y. Zou, and S. Slater. Towards Generalizable Detection of Urgency of Discussion Forum Posts. In M. Feng, T. Käser, and P. Talukdar, editors, \textit{Proceedings of the 16th International Conference on Educational Data Mining}, pages 302--309, Bengaluru, India, July 2023. International Educational Data Mining Society.\\

© 2023 Copyright is held by the author(s). This work is distributed under the Creative Commons Attribution NonCommercial NoDerivatives 4.0 International (CC BY-NC-ND 4.0) license.\\
\url{https://doi.org/10.5281/zenodo.8115790}}

\maketitle

\begin{abstract}
Students who take an online course, such as a MOOC, use the course's discussion forum to ask questions or reach out to instructors when encountering an issue. However, reading and responding to students' questions is difficult to scale because of the time needed to consider each message. As a result, critical issues may be left unresolved, and students may lose the motivation to continue in the course. To help address this problem, we build predictive models that automatically determine the urgency of each forum post, so that these posts can be brought to instructors' attention. This paper goes beyond previous work by predicting not just a binary decision cut-off but a post's level of urgency on a 7-point scale. First, we train and cross-validate several models on an original data set of 3,503 posts from MOOCs at University of Pennsylvania. Second, to determine the generalizability of our models, we test their performance on a separate, previously published data set of 29,604 posts from MOOCs at Stanford University. While the previous work on post urgency used only one data set, we evaluated the prediction across different data sets and courses. The best-performing model was a support vector regressor trained on the Universal Sentence Encoder embeddings of the posts, achieving an RMSE of 1.1 on the training set and 1.4 on the test set. Understanding the urgency of forum posts enables instructors to focus their time more effectively and, as a result, better support student learning.
\end{abstract}

\keywords{educational data mining, learning analytics, text mining, natural language processing, forum post urgency}

\section{Introduction}

In computer-supported learning environments, students often ask questions via email, chat, forum, or other communication media. Responding to these questions is critical for learners’ success since students who do not receive a timely reply may struggle to achieve their learning goals. In a small-scale qualitative study of online learning~\cite{despres2018student}, students who received delayed responses to their questions from the instructor reported lower satisfaction with the course. Another study showed that students who received instructor support through personalized emails performed better on both immediate quizzes and delayed assessments~\cite{kurtz2022impact}.

Massive Open Online Courses (MOOCs) are a prevalent form of computer-supported learning. MOOCs enable many students worldwide to learn at a low cost and in a self-paced environment. However, many factors cause students to drop out of MOOCs, including psychological, social, and personal reasons, as well as time, hidden costs, and course characteristics~\cite{wang2022factors}.

A MOOC’s discussion forum is central to decreasing the risk of student drop-out since it promotes learner engagement with the course. Students use the forum to ask questions, initiate discussions, report problems or errors in the learning materials, interact with peers, or otherwise communicate with the instructor. Andres et al.~\cite{Andres2018} reviewed studies on MOOC completion and discovered that certain behaviors, such as spending above-average time in the forum or posting more often than average, are associated with a higher likelihood of completing the MOOC. Similarly, Crues et al.~\cite{Crues2018} showed that students who read or write forum posts are more likely to persist in the MOOC. At the same time, instructor participation in the forum and interaction with students promotes engagement with the course~\cite{zhang2018role}.

For the reasons above, the timely response of instructors to students’ posts is important. In a study with 89 students, 73 of them preferred if the instructor responded to discussion forum posts within one or two days~\cite{larson2019goldilocks}. However, this is not always feasible. Students’ posts that require an instructor’s response may be unintentionally overlooked due to MOOCs’ scale. Instructors can feel overwhelmed by a large number of posts and often lack time to respond quickly enough or even at all. As a result, issues that students describe in the forum are left unsolved~\cite{Alrajhi2020}, leaving the learners discouraged and frustrated. 

\subsection{Problem Statement}

Since MOOCs tend to have far more students than other computer-supported learning environments, identifying urgent student questions is crucial. We define urgency in discussion forum posts as the degree of how quickly the instructor’s response to the post is needed. Urgency is expressed on an ordinal scale from 1 (not urgent at all) to 7 (extremely urgent). This scale is adopted from the Stanford MOOCPosts data set~\cite{agrawal2014stanford}, arguably the most widely used publicly available data set of MOOC discussion forum posts. It contains 29,604 anonymized, pre-coded posts that have been employed in numerous past studies (see Section 2).

Educational data mining and natural language processing techniques may allow us to automatically categorize forum posts based on their urgency. Our goal is to build models that will perform such categorizations to determine whether a timely response to a post would be valuable. Ultimately, we aim to help instructors decide how to allocate their time where it is needed the most.

Automatically determining the urgency of forum posts is a challenging research problem. Since posts highly vary in content – the students can type almost anything – the data may contain a lot of noise that is not indicative of urgency. In addition, it is difficult to generalize the trained models to other contexts because of linguistic differences caused by different variants of English or by non-native speakers of English, as well as terms that are highly specific to a course topic.

\subsection{Contributions of This Research}

We collected and labeled an original data set of 3,503 forum posts, which we used to train and cross-validate several classification and regression models. From the technical perspective, we tested two different families of features and compared the performance of the regressors, multi-class classifiers, and binary classifiers.

Subsequently, we tested the generalizability of the results by using the independent Stanford MOOCPosts data set~\cite{agrawal2014stanford} of 29,604 forum posts as our holdout test set.

\section{Related Work}

Almatrafi et al.~\cite{ALMATRAFI2018} used the Stanford MOOCPosts data set to extract three families of features: Linguistic Inquiry and Word Count (LIWC) attributes, term frequency, and post metadata. They represented the problem of urgency prediction as binary classification, considering the post not urgent if it had a label below 4, and urgent for 4 and above. The study evaluated five classification approaches: Naive Bayes, Logistic Regression, Random Forest, AdaBoost, and Support Vector Machines. The best-performing model was AdaBoost, able to classify the forum post urgency with the weighted F1-score of 0.88.

Sha et al.~\cite{sha2021hammer} systematically surveyed approaches for classifying MOOC forum posts. They discovered that previous research used two types of features: textual and metadata. Textual features consist of n-grams, post length, term frequency-inverse document frequency (TF-IDF), and others. Metadata features include the number of views of the post, the number of votes, and creation time. Furthermore, the survey compared six algorithms used to construct urgency models from these features, building on the methods by Almatrafi et al.~\cite{ALMATRAFI2018}. Four traditional machine learning (ML) algorithms included Naive Bayes, Logistic Regression, Random Forest, and Support Vector Machines. The best results were yielded by combining textual and metadata features and training a Random Forest model (AUC = 0.89, F1 = 0.89). Two deep learning algorithms examined in the survey were CNN-LSTM and Bi-LSTM. Using the same metrics, these models performed even better than the traditional ones. However, in their follow-up work, Sha et al.~\cite{Sha2022} concluded that deep learning does not necessarily outperform traditional ML approaches overall. The best urgency classifier, again a Random Forest model, achieved an F1-score of 0.90 (AUC was not reported).

Several studies employed the Stanford MOOCPosts data set to train a neural network (NN) for identifying urgent posts. Capuano and Caballé~\cite{Capuano2020} created a 2-layer feed-forward NN on the Bag of Words representation of the posts, reaching an F1-score of 0.80. Alrajhi et al.~\cite{Alrajhi2020} used a deep learning model that combined text data with metadata about posts. They reported an F1-score of 0.95 for predicting non-urgent posts (defined by labels 1–4) and 0.74 for predicting urgent posts (label > 4). Yu et al.~\cite{Yu2021} also transformed the problem into binary classification. They compared three models, the best being a recurrent NN achieving an F1-score of 0.93 on non-urgent posts and 0.70 on urgent posts.

More advanced approaches include those by Guo et al.~\cite{Guo2019}, who proposed an attention-based character-word hybrid NN with semantic and structural information. They achieved much higher F1-scores overall, ranging from 0.88 to 0.92. Khodeir~\cite{Khodeir2021} represented the Stanford MOOCPosts data set using BERT embeddings and trained gated recurrent NNs to predict the posts’ urgency. The best model achieved weighted F1-scores from 0.90 to 0.92.

Previous work used the Stanford MOOCPosts data set to train the models but did not evaluate them on other data. Therefore, the models may overfit to that data set but be ineffective in other contexts. By training models on our own data and testing it on the Stanford MOOCPosts data set, we provide a new perspective within the current body of work in post urgency prediction. We aim to achieve a more generalizable modeling of forum posts’ urgency and provide valuable information for instructors who support large numbers of learners.

In doing so, we also build upon work by Wise et al.~\cite{Wise2017}, who researched techniques for determining which MOOC forum posts are related content-wise. They used the Bag of Words representation of posts and extracted unigrams and bigrams as features. Using a Logistic Regression model, they reached an accuracy between 0.73 and 0.85, depending on the course topic. We use similar methods but for a different purpose.

In designing responses to urgent posts, it is valuable to consider the work by Ntourmas et al.~\cite{Ntourmas2022}, who analyzed how teaching assistants respond to students’ forum posts in two MOOCs. The researchers combined content, linguistic, and social network analysis to discover that teaching assistants mostly provide direct answers. The researchers suggested that this approach does not adequately promote problem-solving. Instead, they argued that more indirect and guiding approaches could be helpful.

\section{Research Methods}

This section describes the data and approaches used to train and evaluate predictive models of forum post urgency.

\subsection{Data Collection and Properties}

We collected posts from students who participated in nine different MOOCs at the University of Pennsylvania (UPenn) from the years 2012 to 2015. The nine MOOCs focused on a broad range of domains (in alphabetical order): accounting, calculus, design, gamification, global trends, modern poetry, mythology, probability, and vaccines. This breadth of covered topics enables us to prevent bias towards certain course topics and support generalization across courses.

To construct the research data set, we started by randomly sampling 500 forum posts for each of the nine courses. Then, we removed posts that:
\begin{itemize}
    \item were in a language other than English
    \item contained only special symbols and characters
    \item contained only math formulas
    \item contained only website links
\end{itemize}

As a result, we ended up with 3,503 forum posts from 2,882 students. This data set included a similar number of posts from each course (between 379 and 399 per course), adding up to the total of 3,503.

Each data point consists of three fields: a unique numerical student ID, the timestamp of the forum post submission, and the post text. All remaining post texts are in the English language, though not all students who wrote them were native speakers of English. The posts contain typos, grammatical errors, and so on, which we did not correct.

\subsection{Data Anonymization}

To preserve student privacy, two human readers manually redacted personally identifiable information in the posts. The removed pieces of text included names of people or places, contact details, and any other information that could be used to determine who a specific poster was. 

Each of the two readers processed roughly half of the post texts from each of the nine courses (195 posts per course per reader on average). The split was selected randomly.

After this anonymization procedure was completed, the data were provided to the research team. To support the replicability of our results, the full data set used in this research can be found at \url{https://github.com/pcla-code/forum-posts-urgency}.

Since we use only de-identified, retrospective data, and the numerical student IDs cannot be traced back to the students’ identity, this research study received a waiver from the university’s institutional review board.

\subsection{Data Labeling}

Three human coders (distinct from those individuals who anonymized the data) manually and independently labeled the 3,503 anonymized post texts. To ensure the approach was unified, they completed coder training and followed a predefined protocol that specified how to assign an urgency label to each post. The protocol is available alongside our research data at \url{https://github.com/pcla-code/forum-posts-urgency}. 

The three coders initially practiced on a completely separate data set of 500 labeled posts with the urgency label hidden. After each coded response, they revealed the correct label and consulted an explanation if they were off by more than 1 point on the scale.

At the end of the training, we computed the inter-rater reliability of each coder within the practice set. Specifically, we calculated continuous (i.e., weighted) Cohen’s Kappa using linear weighting. The three coders achieved the Kappa of 0.57, 0.49, and 0.56, respectively. We note that the weighted values are typically lower than regular Kappa. For instance, weighted Kappa values are lower when there is a relatively large number of categories~\cite{brenner1996dependence}, as is seen in our data sets. They are also lower in cases where, for example, one coder is generally stricter than another (i.e., different means by coder) even though their ordering of cases is identical~\cite{schuster2004note}.

When the coders felt confident in coding accurately, the study coordinator sent them 20 different posts from the separate data set with the urgency label removed. If they coded them accurately, they received a batch of 50 original posts (out of our 3,503 collected) for actual coding. In case a coder was unsure, discrepancies were resolved by discussion.

As stated in Section 1.1, we use the term \textit{urgency} to indicate how fast an instructor should respond to the post. For example, if a post is very urgent, then the instructor or teaching assistant (TA) should respond to it as soon as possible. If a post is not urgent, then the instructor and TA might not have to respond to the post at all. Degrees of urgency were mapped to ordinal scores proposed by Agrawal and Paepcke~\cite{agrawal2014stanford} (and later adopted by related work~\cite{ALMATRAFI2018, Alrajhi2020}) as follows:

\begin{itemize}
    \item 1: No reason to read the post
    \item 2: Not actionable, read if time
    \item 3: Not actionable, may be interesting
    \item 4: Neutral, respond if spare time
    \item 5: Somewhat urgent, good idea to reply, a teaching assistant might suffice
    \item 6: Very urgent: good idea for the instructor to reply
    \item 7: Extremely urgent: instructor definitely needs to reply
\end{itemize}

Example for label 1: \textit{“Hi my name is [REDACTED] and I work in the healthcare industry, looking forward to this course!“}

Example for label 5: \textit{“When will the next quiz be released? I'd like to get a head start on it since I've got some extra time these days.”}

Example for label 7: \textit{“The website is down at the moment, [link] seems down and I'm not able to submit the Midterm. Still have the "Final Submit" button on the page, but it doesn't work. Are the servers congested?”}

Table 1 lists the frequencies of individual urgency labels in the training data across each of the nine courses, as well as their total count. We also detail the frequencies of urgency labels in our test set (see Section 3.6). As the table shows, the frequencies of the labels differ between the training and test set; thus, if our models perform well in this case, they are likely to be robust when predicting data with various distributions.

\begin{table}[t]
\centering
\renewcommand{\arraystretch}{1.2}
\vspace*{-2.7mm} 
\caption{Distribution of training labels in each course. The row \textit{Train} is the sum of all the label frequencies in the individual courses. The row \textit{Test} is the distribution of the labels in the separate test set (rounded up, see Section 3.6).}

\begin{tabular}{l|rrrrrrr}
Course & 1 & 2 & 3 & 4 & 5 & 6 & 7 \\ \hline
Accoun & 199 & 63 & 18 & 53 & 48 & 6 & 0 \\
Calcul & 64 & 167 & 44 & 88 & 31 & 2 & 0 \\
Design & 148 & 114 & 36 & 31 & 35 & 15 & 1 \\
Gamif & 243 & 62 & 15 & 31 & 28 & 0 & 0 \\
Global & 123 & 197 & 21 & 16 & 25 & 5 & 0 \\
Modern & 131 & 214 & 30 & 15 & 7 & 1 & 0 \\
Mythol & 129 & 149 & 59 & 24 & 24 & 5 & 0 \\
Probab & 125 & 115 & 48 & 72 & 31 & 5 & 0 \\
Vaccin & 114 & 139 & 63 & 43 & 21 & 9 & 1 \\ \hline
\textit{Train} & 1276 & 1220 & 334 & 373 & 250 & 48 & 2 \\
\textit{Test} & 3501 & 14997 & 3308 & 3054 & 2259 & 2471 & 14 \\
\end{tabular}
\end{table}

\subsection{Data Automated Pre-Processing}

Before training the models, we performed automated data cleaning and pre-processing that consisted of the following steps in this order:
\begin{itemize}
    \item Converting all text in the posts to lowercase.
    \item Replacing all characters, except the letters of the English alphabet and numbers, with spaces.
    \item Removing duplicate whitespace.
    \item Removing common stopwords in the English language, such as articles and prepositions.
    \item Stemming, that is, automatically reducing different grammatical forms of each word to its root form~\cite{jivani2011comparative}.
\end{itemize}
Each pre-processed post contained 51 words on average (stdev 76, min 1, max 1390).

\subsection{Model Training and Cross-Validation}

The problem of assigning a forum post into one of seven ordered categories corresponds to multi-class ordinal classification or regression (Section 3.5.1). In addition, we also converted the problem to binary classification (Section 3.5.2) to provide a closer comparison with related work.

\subsubsection{Multi-class Classification and Regression}

We hypothesized that regression algorithms would be more suitable for our use case because they can capture the order on the 1–7 scale, which categorical classifiers cannot achieve. We used a total of six classification and regression algorithms:
\begin{itemize}
    \item Random Forest (RF) classifier,
    \item eXtreme Gradient Boosting (XGB),
    \item Linear Regression (LR),
    \item Ordinal Ridge Regression (ORR),
    \item Support Vector Regression (SVR) with a Radial Basis Function (RBF) kernel, and
    \item Neural Network (NN) regressor.
\end{itemize}

We used Python 3.10 and standard implementations of the algorithms in the Scikit-learn module~\cite{Pedregosa2011}, using TensorFlow~\cite{abadi2016tensorflow} and Keras~\cite{chollet2015} for the neural networks. The Python code we wrote to train and evaluate the models is available at \url{https://github.com/pcla-code/forum-posts-urgency}.

All algorithms had default hyperparameter values provided by Scikit-learn. The only exception was the neural network with the following settings discovered experimentally:
\begin{itemize}
    \item Input layer with 128 nodes, 0.85 dropout layer, and ReLU activation function,
    \item One hidden layer with 128 nodes, 0.85 dropout layer, and ReLU activation function,
    \item Output layer with 1 node and ReLU activation function.
\end{itemize}

Each algorithm was evaluated on two families of features: one based on \textit{word counts} (Bag of Words or TF-IDF representations of the forum post texts), the other based on \textit{Universal Sentence Encoder v4} (USE)~\cite{cer2018universal} numerical feature embeddings of the forum post texts.

During model training, we used 10-fold student-level cross-validation in each case. The metrics chosen to measure classification/regression performance were Root Mean Squared Error (RMSE) and Spearman $\rho$ correlation between the predicted and actual values of urgency on the validation set. We chose Spearman instead of Pearson correlation because the urgency labels are ordinal data. The output of the regression algorithms was left as a decimal number, i.e., we did not round it to the nearest whole number.

\subsubsection{Binary Classification}

In addition, we trained separate models for binary classification. Following the precedent from the related work~\cite{Alrajhi2020}, the urgency label was converted to 0 if it was originally between 1–4, and converted to 1 if it was originally larger than 4. We did not adopt the approach of Almatrafi et al.~\cite{ALMATRAFI2018}, who considered a post urgent if it was labeled 4 or above, since based on the scale description defined by Agrawal and Paepcke~\cite{agrawal2014stanford} (see Section 3.3), we do not consider “Neutral” posts to be urgent. (When we tried doing this, it caused only a slight improvement in the model performance.)

Then, we trained RF, XGB, and NN classifier models. The performance evaluation metrics were macro-averaged AUC ROC and weighted F1-score.

\subsection{Model Generalizability Evaluation}

To determine the generalizability of our models, we evaluated them on held-out folds of the training set, then tested them on the Stanford MOOCPosts data set. This data set is completely separate from the training and validation sets and should, therefore, indicate how well our models would perform in different courses and settings.

The test set uses the 1–7 labels but with .5 steps, meaning that some posts can be labeled as 1.5 or 6.5, for example. We did not round these during model training to verify generalizability across both types of labels. However, when labeling our training set, we did not consider .5 labels since the coders felt it added too much granularity. Earlier work did not explicitly differentiate the .5 labels from the integers.

\section{Results and Discussion}

This section details the results from both families of models: one based on word counts and the other on Universal Sentence Encoder. Then, we compare our models with those from related literature.

\subsection{Models with Word Count Features}

These models used the Bag of Words or TF-IDF representations of the forum post texts.

\subsubsection{Multi-class Classification and Regression}

We tested the following combinations of settings and hyperparameters for the word count models on the training and cross-validation set:
\begin{itemize}
    \item Method of feature extraction. TF-IDF performed slightly better than Bag of Words.
    \item Range of n-grams extracted from the data. We tried unigrams, bigrams, and a combination of the two. The best results were obtained when using unigrams only. Models based on bigrams only or those that combined unigrams and bigrams performed worse. In the 3,503 posts, we had 774 unigram and 226 bigram features.
    \item Minimal/maximal allowed document frequency for each term. Here, the best-performing cut-off was to discard the bottom/top 1\% of extreme document frequencies, so the ranges were set to 0.01 and 0.99, respectively. Using this approach made the algorithms run substantially faster, but given the extreme cut-offs, it did not appreciably change the values. Without setting the cut-offs, the training of some models took several hours.
    \item Feature unitization. It either did not impact or slightly worsened the model performance in all cases, so we did not use it.
\end{itemize}

Table 2 summarizes the performance of all models. Support vector regression performed best overall on the training and cross-validation set in terms of both metrics: RMSE and Spearman $\rho$ correlation. It also outperformed the other approaches on the separate test set.

\begin{table}[t]
\centering
\renewcommand{\arraystretch}{1.2}
\vspace*{-2.7mm} 
\caption{Performance of multi-class classification/regression models on the training set of 3,503 posts (UPenn) and the test set of 29,604 posts (Stanford). Features: word counts.}

\begin{tabular}{l|rr|rr}
 & \multicolumn{2}{c|}{Training and} & \multicolumn{2}{c}{Different university} \\[-1mm]
 & \multicolumn{2}{c|}{cross-validation set} & \multicolumn{2}{c}{test set} \\ \hline
Model & RMSE & $\rho$ & RMSE & $\rho$ \\ \hline
RF & 1.3550 & 0.4258 & 1.7781 & 0.2676 \\
XGB & 1.3338 & 0.4326 & 1.7419 & 0.3086 \\
LR & 1.2385 & 0.4419 & (large) & 0.3432 \\
ORR & 1.1501 & 0.4750 & 1.4229 & 0.3484 \\
NN & 1.1269 & 0.4897 & 1.4395 & 0.3746 \\
SVR & 1.0946 & 0.5503 & 1.4138 & 0.3982
\end{tabular}
\end{table}

\begin{figure}
\Description{Plot showing the average predicted labels, plus or minus the standard deviation, for SVR with word count features.}
\centering
\includegraphics[width=\columnwidth]{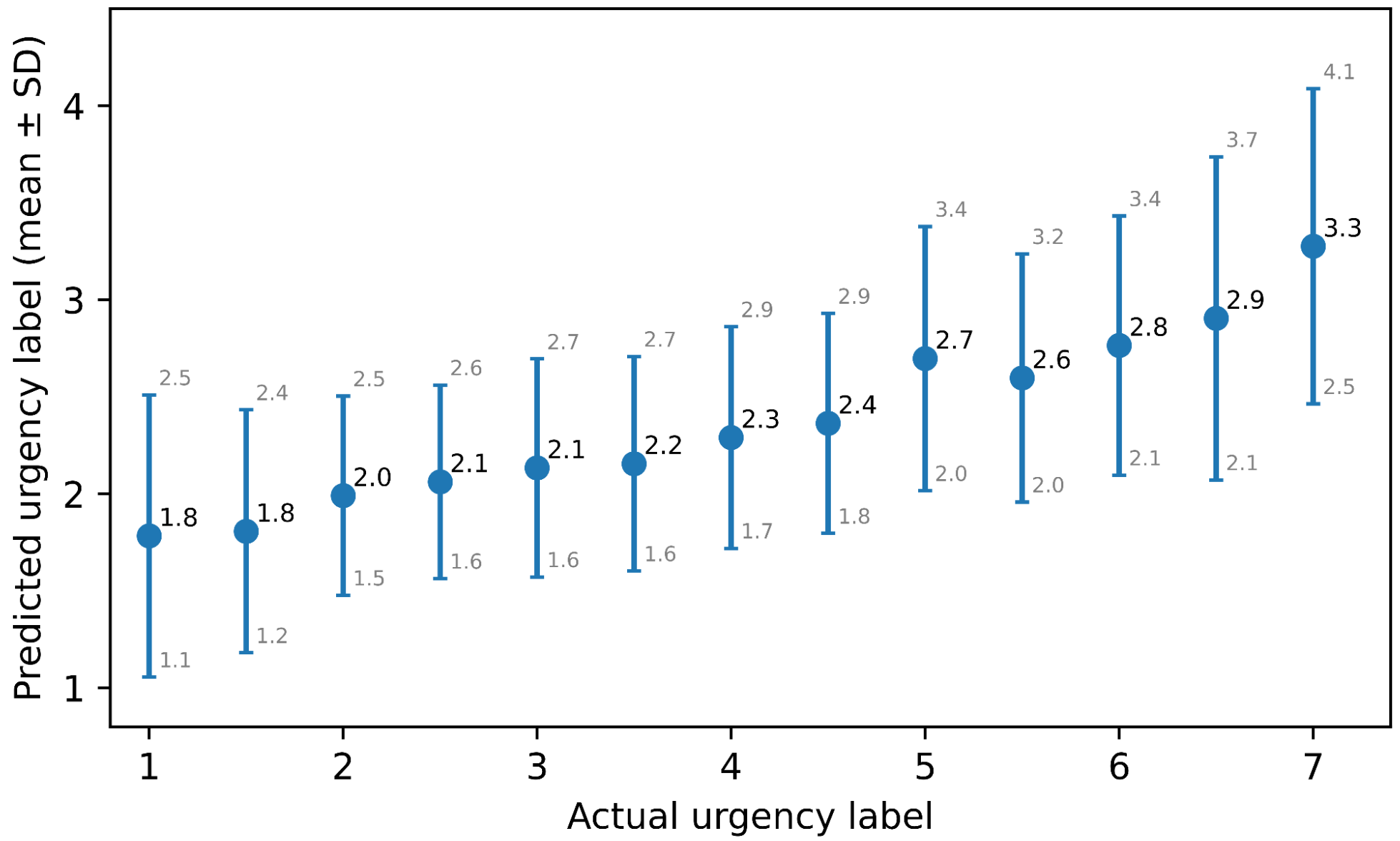}
\caption{Prediction results of the best performing model (SVR) on the separate test set using the word count features.}
\end{figure}

Figure 1 shows the predictions of the best model on the test set. Most urgency labels are under-predicted, but they are still predicted in the increasing order of urgency, which demonstrates that the model is detecting the ranking.

After SVR, other regressors followed, with neural networks being the second best. Overall, the classifier models performed more poorly than the regression models. We expected this result since the urgency classes are ordinal, and the categorical classifiers cannot capture their ordering.

\subsubsection{Binary Classification}

Table 3 summarizes the performance of all models. The NN outperformed the remaining two classifiers, though the differences in AUC are more visible than for F1-score compared to XGBoost. Although the fit of RF and NN is non-deterministic, the results did not change substantially when we re-ran the model training multiple times.

When considering the prediction of non-urgent posts only, all models achieved a very high F1-score between 0.9512 (NN) and 0.9589 (RF) on the training set, and 0.8924 (NN) to 0.8971 (XGBoost) on the test set.

For the urgent posts only, the predictive power was much lower: between 0.1841 (RF) and 0.4168 (NN) on the training set, and 0.0025 (RF) to 0.2761 (NN) on the test set.

Due to the imbalance in favor of the non-urgent class, experimenting with decision cut-offs lower than the default 50\% visibly improved the RF and XGBoost models' AUC (up to 0.7771) but improved the F1-score only slightly. The best results were achieved for decision thresholds of 10 or 15\%.

\begin{table}[t]
\centering
\renewcommand{\arraystretch}{1.2}
\vspace*{-2.7mm} 
\caption{Performance of binary classification models on the training set of 3,503 posts (UPenn) and the test set of 29,604 posts (Stanford). Features: word counts.}

\begin{tabular}{l|rr|rr}
 & \multicolumn{2}{c|}{Training and} & \multicolumn{2}{c}{Different university} \\[-1mm]
 & \multicolumn{2}{c|}{cross-validation set} & \multicolumn{2}{c}{test set} \\ \hline
Model & AUC & F1 & AUC & F1 \\ \hline
RF & 0.5522 & 0.8926 & 0.5005 & 0.7263 \\
XGB & 0.6178 & 0.9053 & 0.5412 & 0.7590 \\
NN & 0.6687 & 0.9055 & 0.5735 & 0.7759
\end{tabular}
\vspace*{-3.75mm}
\end{table}
\FloatBarrier

\subsection{Models with Feature Embeddings Using the Universal Sentence Encoder (USE)}

\subsubsection{Multi-class Classification and Regression}

Table 4 summarizes the performance of all models. Again, SVR performed best on the training set, followed by NN. After that, other regressors and classifiers followed in the same order as with the word-count-based models. However, for the test set, while SVR still obtained the best $\rho$, it had slightly worse RMSE than the other three regressors. Overall, the model quality was better for USE than for TF-IDF.

Figure 2 shows the predictions made by the best model on the test set, with the trend being similar to Figure 1.

\vspace*{-3.25mm}

\begin{table}[!h]
\centering
\renewcommand{\arraystretch}{1.2}
\caption{Performance of multi-class classification/regression models on the training set of 3,503 posts (UPenn) and the test set of 29,604 posts (Stanford). Features: USE embeddings.}

\begin{tabular}{l|rr|rr}
 & \multicolumn{2}{c|}{Training and} & \multicolumn{2}{c}{Different university} \\[-1mm]
 & \multicolumn{2}{c|}{cross-validation set} & \multicolumn{2}{c}{test set} \\ \hline
Model & RMSE & $\rho$ & RMSE & $\rho$ \\ \hline
RF & 1.4707 & 0.3452 & 1.8995 & 0.2723 \\
XGB & 1.3569 & 0.4418 & 1.7753 & 0.3145 \\
LR & 1.1758 & 0.4717 & 1.3953 & 0.3882 \\
ORR & 1.1448 & 0.4983 & 1.3723 & 0.3964 \\
NN & 1.1045 & 0.5361 & 1.3988 & 0.4202 \\
SVR & 1.0956 & 0.5716 & 1.4065 & 0.4283
\end{tabular}
\end{table}

\begin{figure}[!h]
\Description{Plot showing the average predicted labels, plus or minus the standard deviation, for SVR with USE features.}
\centering
\includegraphics[width=\columnwidth]{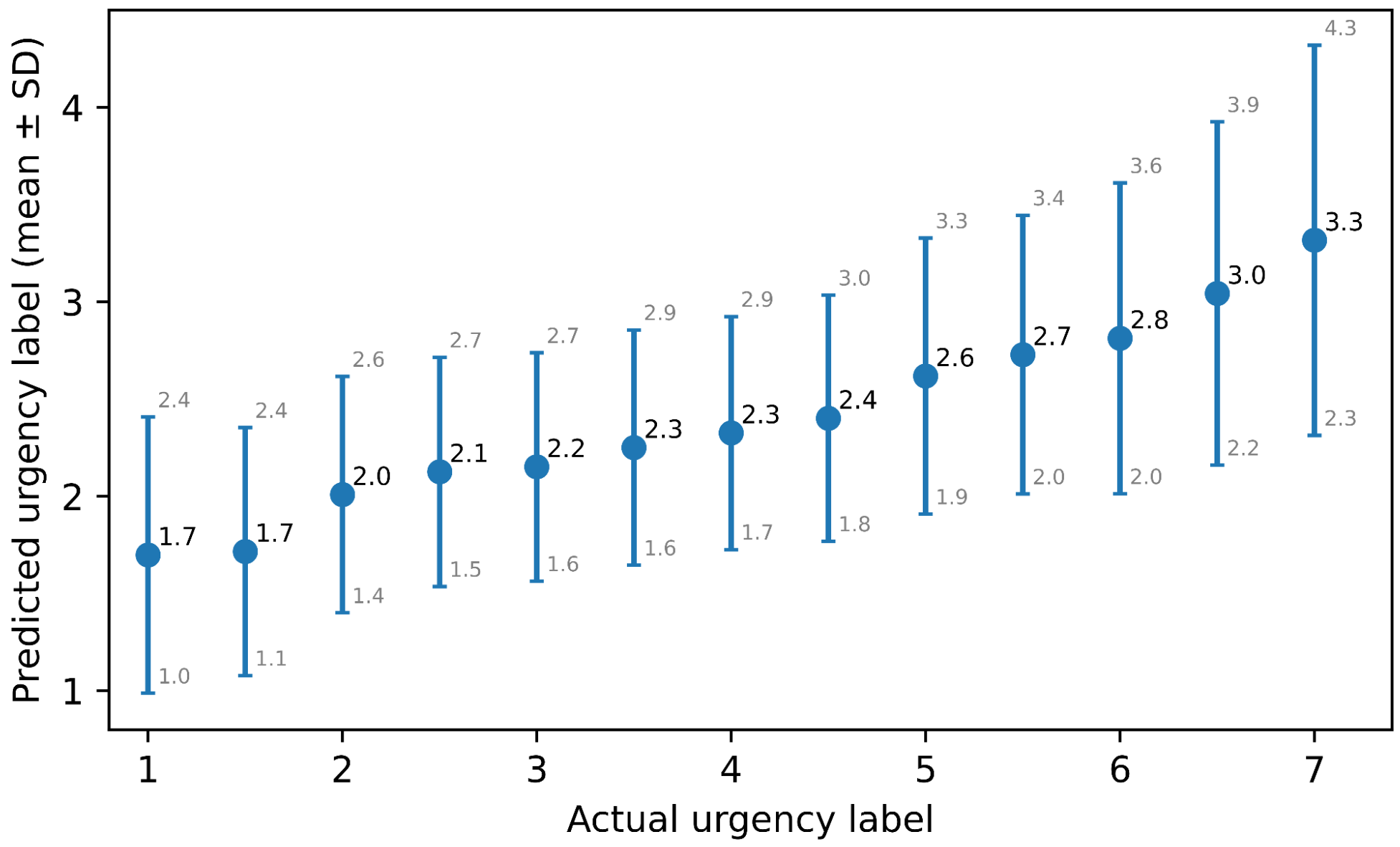}
\caption{Prediction results of the best performing model (SVR) on the separate test set using the USE features.}
\end{figure}

\FloatBarrier

\subsubsection{Binary Classification}

Table 5 summarizes the performance of all models. Compared to using the TF-IDF features, the results are surprisingly slightly worse, even though the differences are minimal in some cases. The overall order of models is preserved -- again, the NN outperformed the other two models.

\begin{table}[t]
\centering
\renewcommand{\arraystretch}{1.2}
\vspace*{-2.7mm} 
\caption{Performance of binary classification models on the training set of 3,503 posts (UPenn) and the test set of 29,604 posts (Stanford). Features: USE embeddings.}

\begin{tabular}{l|rr|rr}
 & \multicolumn{2}{c|}{Training and} & \multicolumn{2}{c}{Different university} \\[-1mm]
 & \multicolumn{2}{c|}{cross-validation set} & \multicolumn{2}{c}{test set} \\ \hline
Model & AUC & F1 & AUC & F1 \\ \hline
RF & 0.5094 & 0.8774 & 0.5002 & 0.7260 \\
XGB & 0.5863 & 0.9020 & 0.5246 & 0.7470 \\
NN & 0.6409 & 0.9054 & 0.5684 & 0.7760
\end{tabular}
\end{table}

As previously, we observed similar imbalances in F1-scores when predicting non-urgent and urgent posts separately. For non-urgent posts, all models achieved a high F1-score between 0.9544 (NN) and 0.9597 (XGBoost) on the training set, and 0.8954 (RF) to 0.8974 (XGBoost) on the test set.

For predicting the urgent posts only, the predictive power is much lower: between 0.0366 (RF) and 0.3799 (NN) on the training set, and 0.0007 (RF) to 0.2563 (NN) on the test set. Again, the respective performance of the individual classifiers corresponds to the case with word count features.

As expected, decreasing the decision cut-off below 50\% again substantially improved the overall model performance. The best results were again achieved for decision thresholds of 10 or 15\%.

\subsection{Comparison with the Results Published in Previous Literature}

We now compare our results with the binary classification models reported in Section 2, which were trained on the Stanford MOOCPosts data set. We cannot compare our multi-class classification and regression analyses to past work since it treated this problem only as binary classification.

Almatrafi et al.~\cite{ALMATRAFI2018} and Sha et al.~\cite{sha2021hammer} slightly differed from our approach in using the label 4 as the cut-off for post urgency, as opposed to 4.5. The best model by Almatrafi et al.~\cite{ALMATRAFI2018}, an AdaBoost classifier, achieved a weighted F1-score of 0.88. Our binary classifiers slightly outperformed this model, even though we used fewer types of features. This indicates that combining features from various sources does not necessarily improve model quality. Sha et al.~\cite{sha2021hammer} reported a RF model that scored F1 = 0.89 and AUC = 0.89. While we achieved similar F1-scores, our AUC was much lower. This could have been caused by the smaller training set, in which the class imbalance had a larger effect.

The NN approaches by Capuano and Caballé~\cite{Capuano2020}, Guo et al.~\cite{Guo2019}, and Khodeir~\cite{Khodeir2021} reported F1-scores ranging from 0.80 to 0.92. Even though our NN models were much simpler and trained on a smaller data set, they achieved a similarly high F1 of 0.91.

Finally, Alrajhi et al.~\cite{Alrajhi2020} and Yu et al.~\cite{Yu2021} reported the model performance separately for non-urgent and urgent posts. When considering non-urgent posts only, they reached F1-scores of 0.95 and 0.93, respectively. Our best-performing model on this task achieved F1 = 0.96 on the training set (RF, word count features) and 0.90 on the test set (XGBoost, USE features). When considering urgent posts only, they reported F1-scores of 0.74 and 0.70. Here, our models scored much worse, 0.42 on the training set and 0.28 on the test set (both approaches used NN on the word count features). The AUC scores were not reported in this case.

Overall, we achieved comparable or even slightly better performance in most cases. In addition, we evaluated the models for multi-class classification and regression, which the previous work did not consider.

We could not fully replicate past work because the feature set and the code used to produce the previous results were unavailable. This prevented us from testing the prior work on our data set, which would have helped to establish the generalizability of those earlier approaches.

\subsection{Opportunities for Future Work}

In future work, the urgency rating of forum posts can also be treated as a ranking problem. Using an ML algorithm, posts can be sorted from the most to the least urgent instead of classifying them as high or low priority. Even among the posts with the same urgency level, some messages should be addressed first. Therefore, reframing the problem to ranking learning would lead to a different model that suggests the most urgent post to address instead of estimating the level of urgency. Our current approach shows that regardless of the regression outputs, regressor models such as the SVR correctly estimate a higher urgency for more urgent posts. For this reason, ML models could show promising results for sorting the posts based on their urgency.

In addition, the post labeling scale could be improved, perhaps by simplifying it to fewer categories. In this study, we adopted the scale from previous work~\cite{agrawal2014stanford}, used additionally in~\cite{ALMATRAFI2018, Alrajhi2020} in order to be able to study the generalizability of findings across data sets. Finally, experimenting with over- or undersampling of the training set using algorithms such as SMOTE might improve model performance for certain labels.

To ensure even a higher degree of generalizability, future research could validate the models on data from different populations than those employed in our paper.

\section{Conclusion}

Responding to students’ concerns or misunderstandings is vital to support students’ learning in both traditional and MOOC courses. Since instructors cannot read all forum posts in large courses, selecting the posts that urgently require intervention helps focus instructors’ attention where needed.

The presented research aims to automatically determine the urgency of forum posts. We used two separate data sets with different distributions and different approaches to the urgency scale (using .5 values or not) to support generalizability. Support vector regression models showed the highest performance in almost all aspects and cases. The best model from both categories of features (word count or numerical embeddings) performed similarly, with Universal Sentence Encoder embeddings being slightly better.

The results of this work can contribute to supporting learners and improving their learning outcomes by providing feedback to instructors and staff managing courses with large enrollment. The model quality has implications for practical use. Based on the RMSE values, it is unlikely that a highly urgent post will be labeled non-urgent and vice versa. From a practical perspective, implementing the urgency rating into MOOC platforms or large courses would help instructors, for example, by providing automated notification on posts with high urgency. In this case, however, students should not be aware of the inner workings of such a system. This is to prevent abuse by writing words with certain phrases to trigger instructor notifications.

\section{Acknowledgments}

This research was supported by the National Science Foundation (NSF) (NSF-OAC\#1931419). Any opinions, findings and conclusions, or recommendations expressed in this paper are those of the authors and do not necessarily reflect the views of the NSF.

We also thank Nathan Levin and Xiaodan Yu for their work in preparing and labeling the research data set.

\bibliographystyle{abbrv}
\bibliography{references}  

\balancecolumns

\end{document}